\documentclass[11pt]{article} 
\usepackage{rldm,palatino}
\usepackage{graphicx}
\usepackage{amsmath}
\usepackage{amssymb}
\usepackage{caption}
\usepackage{subcaption}
\usepackage{comment}
\usepackage{verbatim}
\usepackage[style=ieee]{biblatex}

\newcommand*\samethanks[1][\value{footnote}]{\footnotemark[#1]}
\usepackage[bb=dsserif]{mathalpha}
\usepackage{wrapfig}

\bibliography{bib}

\title{Temporally Extended Successor Representations}

\author{
Matthew J.~Sargent\thanks{Corresponding Author. \texttt{https://github.com/mjsargent} } \\
Department of Computer Science\\
University College London\\
\texttt{m.sargent@cs.ucl.ac.uk} \\
\And
Peter J.~Bentley \\
Department of Computer Science \\
University College London \\
\texttt{p.bentley@cs.ucl.ac.uk} \\
\AND
Caswell Barry\thanks{Equal Contribution} \\
Department of Cell and Developmental Biology \\
University College London \\
\texttt{caswell.barry@ucl.ac.uk} \\
\And
William de Cothi\samethanks \\
Department of Cell and Developmental Biology \\
University College London \\
\texttt{w.decothi@ucl.ac.uk} \\
}

\begin{document}
\
\addtolength{\floatsep}{-9mm}
\addtolength{\belowcaptionskip}{-2.1mm}
\addtolength{\parskip}{-1mm}

\renewcommand\floatpagefraction{.9}
\renewcommand\topfraction{.9}
\renewcommand\bottomfraction{.93}
\renewcommand\textfraction{.1}   
\setcounter{totalnumber}{50}
\setcounter{topnumber}{50}
\setcounter{bottomnumber}{50}
\maketitle

\begin{abstract}    
We present a temporally extended variation of the successor representation, which we term \emph{t-SR}. t-SR captures the expected state transition dynamics of temporally extended actions by constructing successor representations over primitive action repeats. This form of temporal abstraction does not learn a top-down hierarchy of pertinent task structures, but rather a bottom-up composition of coupled actions and action repetitions. This lessens the amount of decisions required in control without learning a hierarchical policy. As such, t-SR  directly considers the time horizon of temporally extended action sequences without the need for predefined or domain-specific options. We show that in environments with dynamic reward structure, t-SR is able to leverage both the flexibility of the successor representation and the abstraction afforded by temporally extended actions. Thus, in a series of sparsely rewarded gridworld environments, t-SR optimally adapts learnt policies far faster than comparable value-based, model-free reinforcement learning methods. We also show that the manner in which t-SR learns to solve these tasks requires the learnt policy to be sampled consistently less often than non-temporally extended policies.
\end{abstract}

\keywords{
reinforcement learning; temporal abstraction; successor representation
}

\acknowledgements{Matthew J.~Sargent was funded by an EPSRC Doctoral Training Partnership Research Studentship (EP/R513143/1). Caswell Barry was funded by a Wellcome Senior Research Fellowship (212281/Z/18/Z). 
Matthew J.~Sargent would like to thank Augustine N.~Mavor-Parker for his assistance in data visualisation.}

\startmain 

\section{Introduction}
In reinforcement learning (RL), agents must solve sequential decision making tasks to maximise some notion of expected cumulative reward, known as the value function. Often the solutions to these tasks share elements in terms of their structure - different tasks can be decomposed into sequences of commonly shared sub-tasks. These sub-tasks link the timescales of primitive actions that take place at every step, and that of the broader abstract task, facilitating planning over these extended timescales. As such, \emph{temporal abstraction} is the process by which an agent learns the temporal structure of a task in a way that can both reduce cognitive load and improve generalisation across tasks with shared structure.

The canonical approach to address this form of abstraction is the Options framework \cite{sutton1999between}. Agents using Options seek to learn a set of policies associated with distinct sub-tasks, as well as the initiation and termination conditions for these policies. While this successfully quantifies the temporal abstraction problem, without predefined handcrafted domain-specific Options this process can be more complex than learning simple value functions \cite{pateria2021hierarchical}.

Rather than learning a specific policy associated with each sub-task, an alternative approach to temporal abstraction is to directly consider the time horizon of the actions taken in tandem with the policy for selecting them. This approach of temporally extending primitive actions by repeating them has been successfully applied to model-free RL; an agent's action space can be augmented with temporally extended actions \cite{lakshminarayanan2017dynamic}, or an additional policy can be learnt over the space of possible action repeats \cite{DBLP:conf/iclr/SharmaLR17, biedenkapp-icml21}. This is in contrast to the typical usage of fixed action repetition \cite{Braylan2015FrameSI, hessel2019inductive} in popular RL environments such as the Atari Arcade Learning Environment \cite{bellemare2013}, where action repetition is a fixed property of the environment rather than a process determined by the agent.

Given the success of temporally extended model-free methods, we set out to investigate whether temporally extending the successor representation (SR) could improve generalisation in a similar way. The SR linearly decomposes the problem of learning the value function into independently learning two components: a matrix of expected discounted state occupancies, and a vector of state-dependant expected rewards \cite{dayan1993improving}. Thus, the SR is able to improve upon the generalisation of model-free methods in tasks where the transition or reward structures changes \cite{russek2017predictive}. 

We show that a temporally extended variant of the SR, which we term \emph{t-SR}, performs comparably to temporally-extended model-free approaches in static, unchanging environments. Further, t-SR exhibits dramatically improved performance over temporally-extended model-free methods when in environments with dynamic reward structure - specifically we assess reward revaluation. 
\section{Preliminaries}
\textbf{Learnt Action Repetition}
In RL, a given task is characterised by a Markov Decision Process (MDP)  $\mathcal{M} = \langle \mathcal{S}, \mathcal{A}, \mathcal{P}, \mathcal{R} \rangle $. Here, the state space $\mathcal{S}$ contains all the states the agent can occupy and the action space $\mathcal{A}$ describes the allowable actions in each state. The transition probabilities $\mathcal{P}$ then define the probability $p(s'|s,a) \dot = \text{Pr}(S_t = s' | S_{t-1} = s, A_{t-1} = a)$ of transitioning from state $s$ to state $s'$ having taken action $a$, with $\mathcal{R}$ being the rewards $r(s,a) \dot = \mathbb{E}\left[R_t | S_{t-1} = s, A_{t-1} = a \right]$ that the agent receives after doing so. The agent is tasked with learning a policy $\pi (a | s)$ which specifies the probability of choosing each action given the current state. Specifically, for a given finite episodic MDP, the agent's task is to learn an optimal policy $\pi^{*}(a | s)$ that maximises expected cumulative discounted reward across an episode.

Conventionally, these actions are single-step primitive actions - the agent resamples its policy at $s'$ to determine the next action. Recent work has extended this action selection policy with a \emph{holding-time} (or \emph{skip}) for the action \cite{DBLP:conf/iclr/SharmaLR17,biedenkapp-icml21}. This is the number of times the chosen action will be repeated before the action selection policy is queried again. Agents are therefore equipped with both an action selection policy $\pi^a : \mathcal{S} \rightarrow \mathcal{A}$, and an action repetition policy $\pi^j : \mathcal{S} \rightarrow \mathcal{J}$. Here we follow the notation of \cite{biedenkapp-icml21} and let $j$ be an action repetition number, with $\mathcal{J}$ being the set of allowable action repetitions.

The action repetition policy $\pi^j$ can be learnt using the same methods used to learn a standard policy. Here, we use the $Q$-learning style update rule for learnt action repetitions derived in \cite{biedenkapp-icml21} due to its relevance to constructing SRs \cite{watkins1992q,dayan1993improving}, but methods for learning repetition policies through policy gradients also exist \cite{DBLP:conf/iclr/SharmaLR17}. An optimal $\pi^{j*}$ can be obtained by greedily choosing from the optimal $Q^{\pi j*}(s, j |a)$, with the action-repetition values drawn from transitions made in the environment according to:
\begin{equation}
    Q^{\pi j}(s, j | a) = \mathbb{E}^\pi \left[ \sum_{k=0}^{j-1} \gamma^k r_{t+k} + \gamma^j Q^\pi (s_{t+j}, a*_{t_j}) | s_t = s, a, j) \right]
\end{equation}
where $a*$ = $\text{argmax}_a \left[Q^\pi(s_{t+j}, a) \right]$ and $\gamma \in [0,1]$ represents a temporal discount factor that exponentially downweights distal rewards.

\textbf{Successor Representations} The SR was originally introduced as a method to improve generalisation in model-free RL \cite{dayan1993improving}, which it achieves by decomposing the value function into expected state occupancies and state-rewards. The SR itself, is therefore defined as the expected discounted future occupancy of state $s'$ along a trajectory initiated in state $s$ by taking action $a$: 
\begin{equation}
    M^\pi(s,a,s') = \mathbb{E}^\pi \left[ \sum_{t=0}^\infty  \gamma^t\mathbb{1}(s = s') | s_0 = s, a_0 = a \right] 
\end{equation}
where $\mathbb{1}$ is the indicator function. State-action values for a given policy can then be obtained by linearly combining $M^\pi$ with the rewards associated with each state, which can be learnt separately via supervised methods. Thus, the SR allows for rapid reward revaluation by reusing $M^\pi$ - the learnt expected state dynamics for a given policy. 

\section{Temporally Extended Successor Representations}
\begin{figure}
    \centering
    \makebox[\linewidth][c]{
    \begin{subfigure}{0.22\textwidth}
    \centering
    \includegraphics[width=0.95\textwidth]{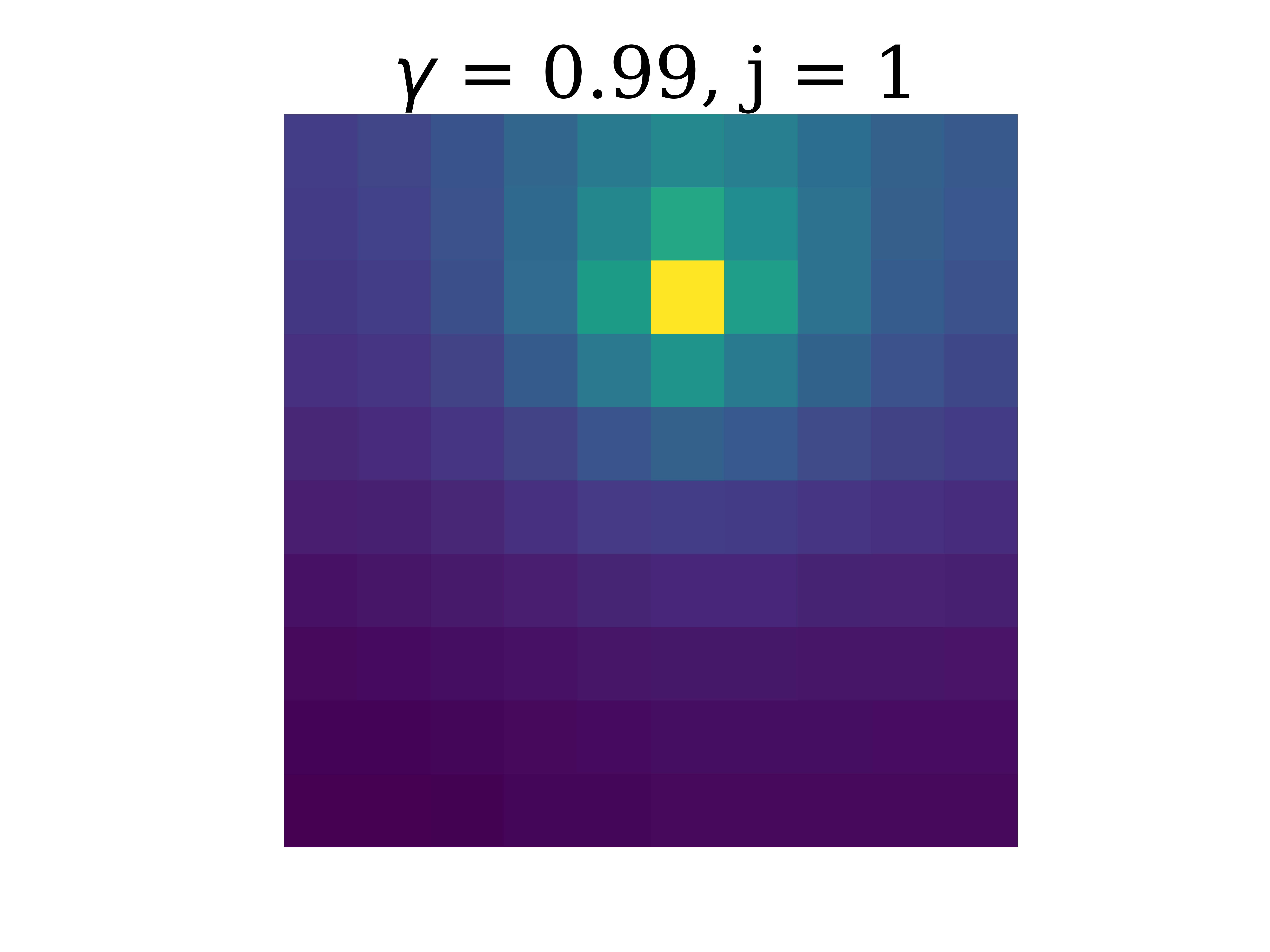}
    \label{fig:gamma099}
    \end{subfigure}
    
    \begin{subfigure}{0.22\textwidth}
    \centering
    \includegraphics[width=0.95\textwidth]{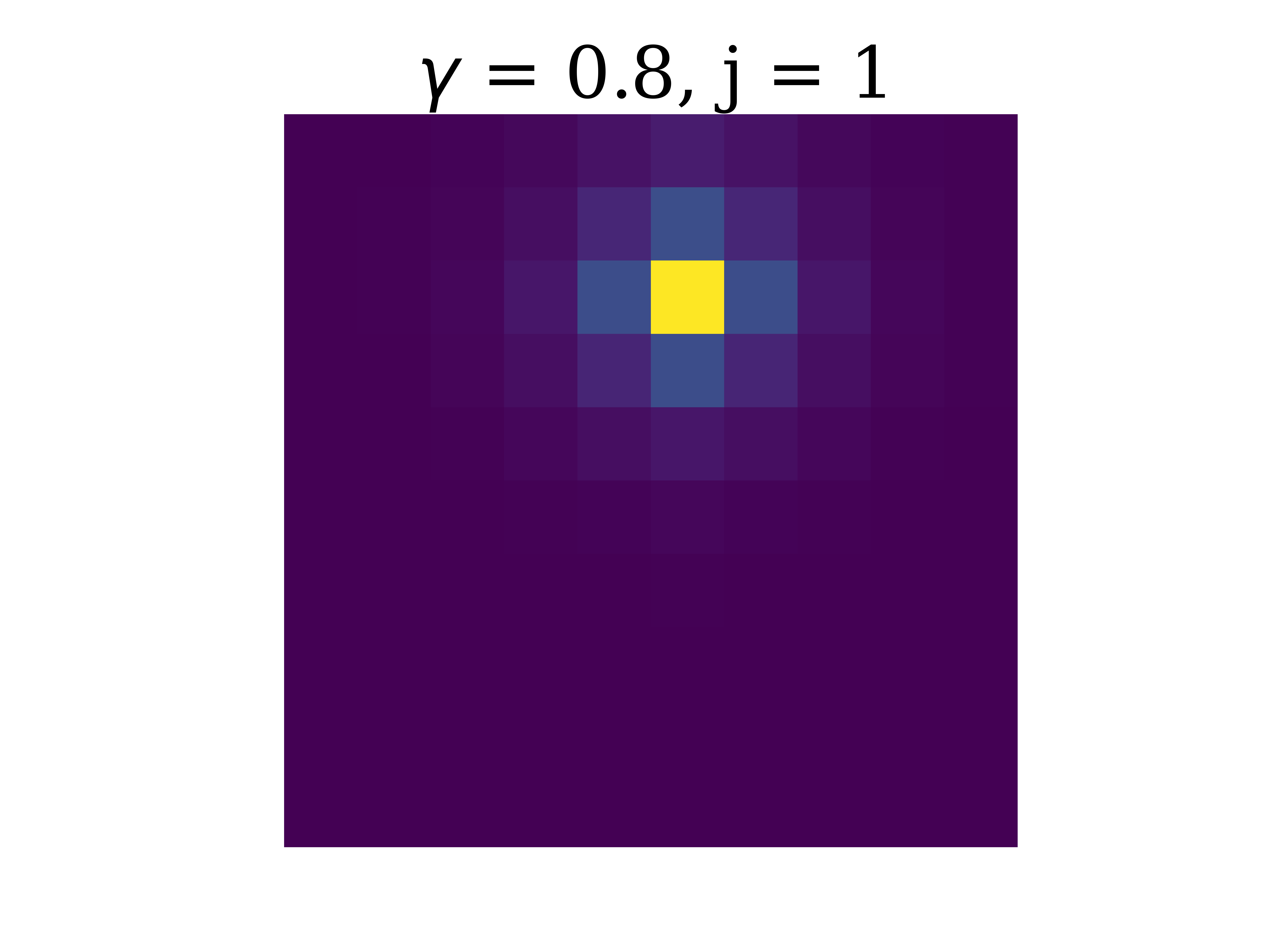}
    \label{fig:gamma08}
    \end{subfigure}
    
    \begin{subfigure}{0.22\textwidth}
    \centering
    \includegraphics[width=0.95\textwidth]{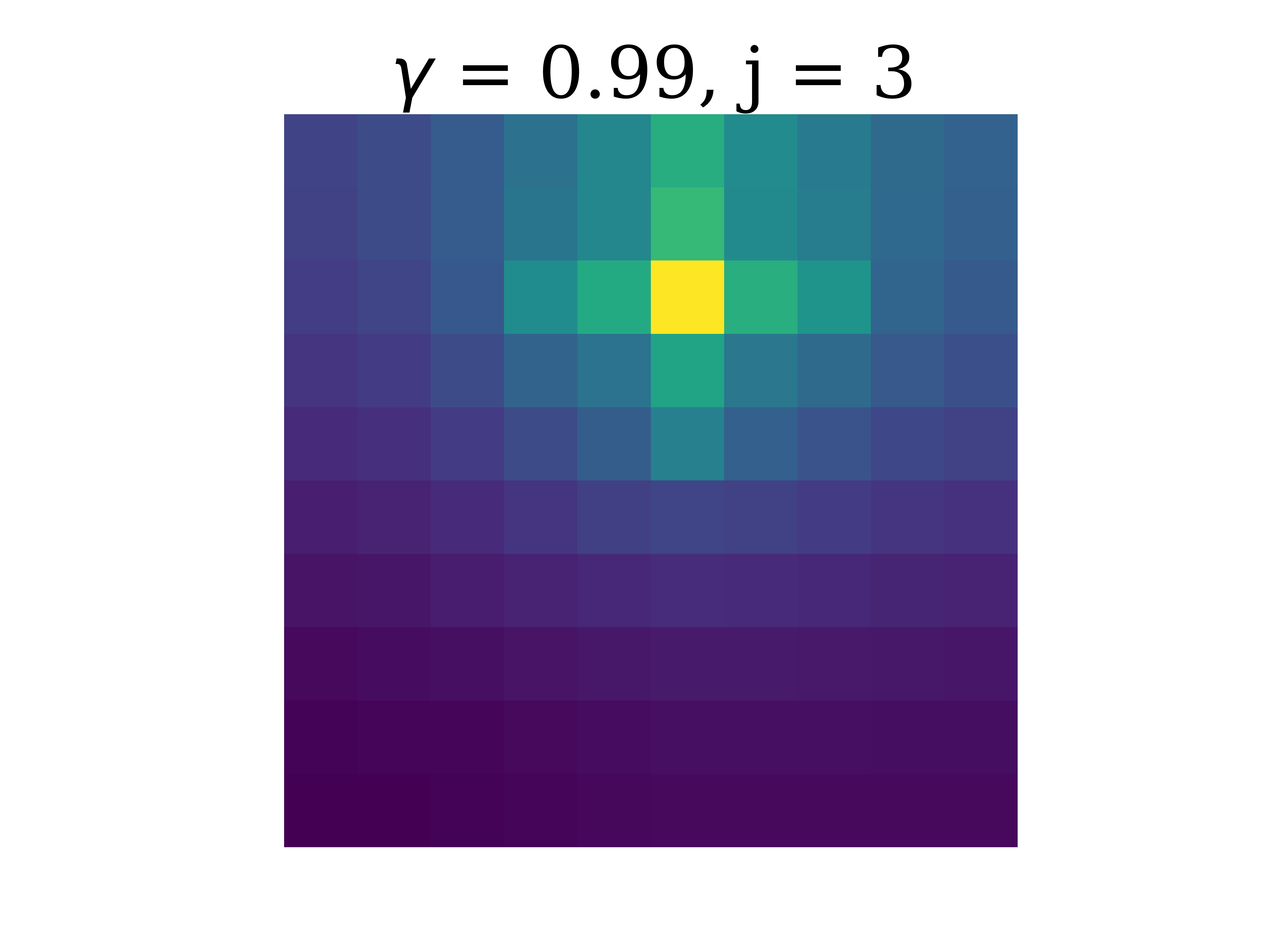}
    \label{fig:j2}
    \end{subfigure}
    \begin{subfigure}{0.22\textwidth}
    \centering
    \includegraphics[width=0.95\textwidth]{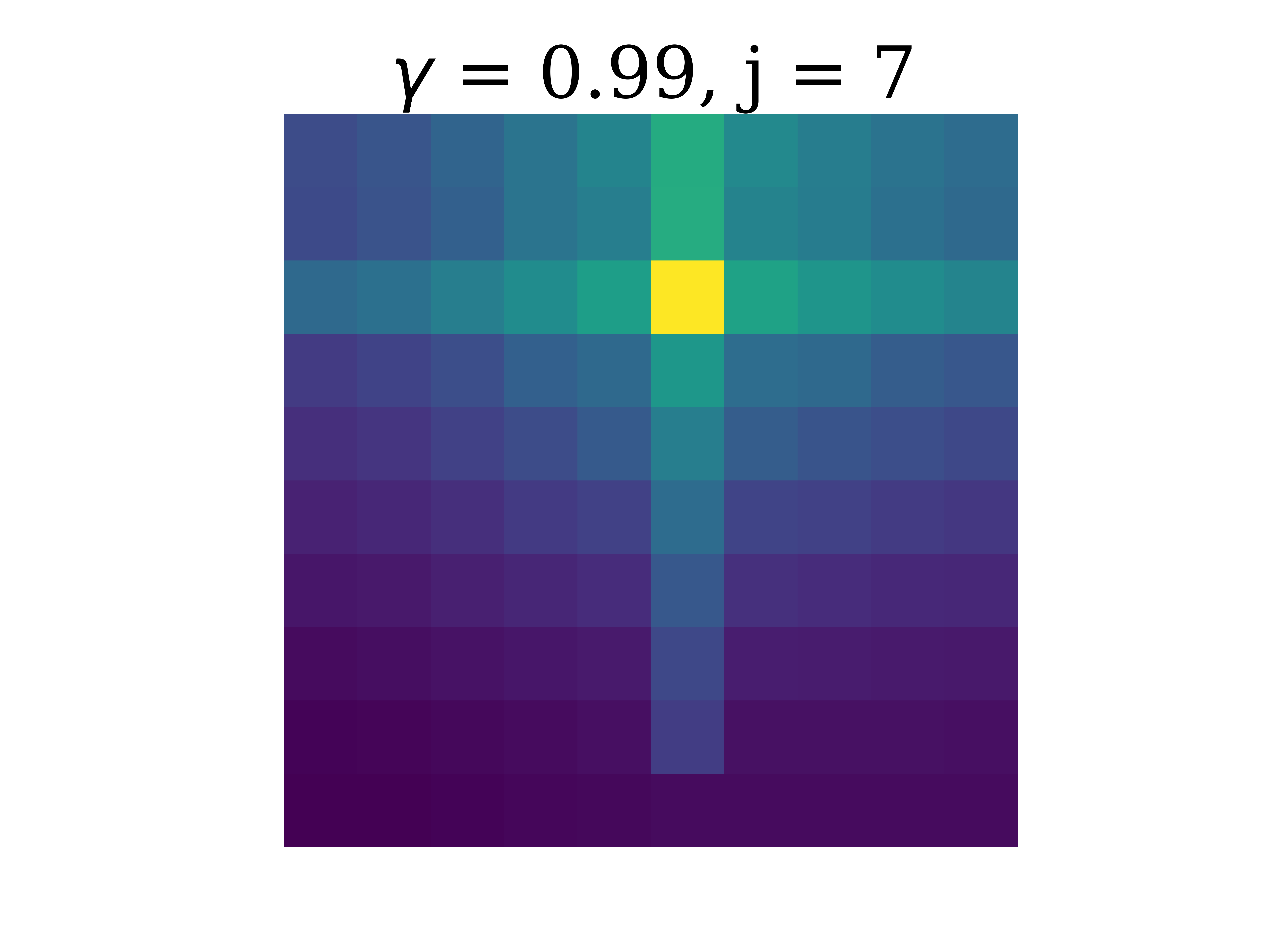}
    \label{fig:j7}
    \end{subfigure}
    }
    \caption{Comparison of SRs for different $\gamma$ and action repetition $j$. These SRs are taken under a random walk policy in a 10x10 gridworld with an action space consisting of the cardinal directions. Each figure consists of a single column of the SRs, projected over the gridworld. Projecting the column over the grid therefore shows the expected likelihood of having started in a state and moving to the state $i$ under the random policy. These figures show the different temporal effects of changing $\gamma$ and varying the action repeat $j$. Increasing action repetition leads to representations that are heavily skewed in the directions of actions. }
    \label{fig:place_fields}
\end{figure}

Expected dynamics can be temporally extended in the same fashion as value functions. Consider the matrix formulation of the SR: $\mathbf{M^\pi} = \sum^\infty_{t=0}\left(\gamma \mathbf{T}^\pi \right)^t$, where $\mathbf{T}^\pi$ is the expected one-step transition function between states following policy $\pi$. To specify a SR that considers action repetition, let $\mathbf{T}_a$ be the expected transition conditioned on action $a$ such that $\mathbf{T}^\pi = \mathbb{E}^\pi(\mathbf{T}_a) $. The temporally extended successor representation (t-SR) for an action repetition of $j$ can then be obtained by: 
\begin{equation}
    \mathbf{M}^{\pi j} = \mathbb{E}^\pi \left[\sum_{k=0}^{j} \left( \gamma \mathbf{T}_a \right)^k \right] + {\mathbb{E}^{\pi}\left[(\gamma \mathbf{T}_a)^j \right]}  \left(\mathbb{I} - \gamma \mathbf{T}^{\pi} \right)^{-1} 
\end{equation}
where the right most expression is derived from the converging infinite series (Neumann series) of the standard SR. Compared to the SR, these representations skew according to the action space, reflecting the extended transitions instigated by repetition of the same action - this leads to a markedly different representation of distal state than altering the discount factor $\gamma$ \cite{Momennejad2018PredictingTF} (Fig \ref{fig:place_fields}).

To use the SR for determining action repetition, an action is first sampled from $\pi^a$ such that $M(s,a,s',j)$ now has two of its arguments specified: $s$ and $a$. The action repeat is then the $\max_j$ of $M(s, a, s', j)W$, where $W$ is the learnt reward vector. The baseline SR is learnt through temporal difference learning with state occupancies acting as cummulant rewards, while the reward weights $W$ are learnt through online supervised learning based on the expected rewards and actual rewards obtained. Thus, the t-SR can be learnt using using a typical temporal difference learning rule:
\begin{gather}
        M^{\pi j}(s,a,s', j) \xleftarrow{\alpha} M^{\pi j}(s,a,s', j) + \mathbb{E}^{\pi}\left[ \sum_{k=0}^{j-1} \gamma ^k\mathbb{1}(s_k = s) + \gamma^j M^{\pi}(s_{t+j}, a*, s') - M^{\pi j}(s,a,s',j)\right]
\end{gather}

\section{Experiments}
\begin{figure}
    \centering
    \makebox[\linewidth][c]{
    \begin{subfigure}{0.14\textwidth}
    \centering
    \includegraphics[width=0.95\textwidth]{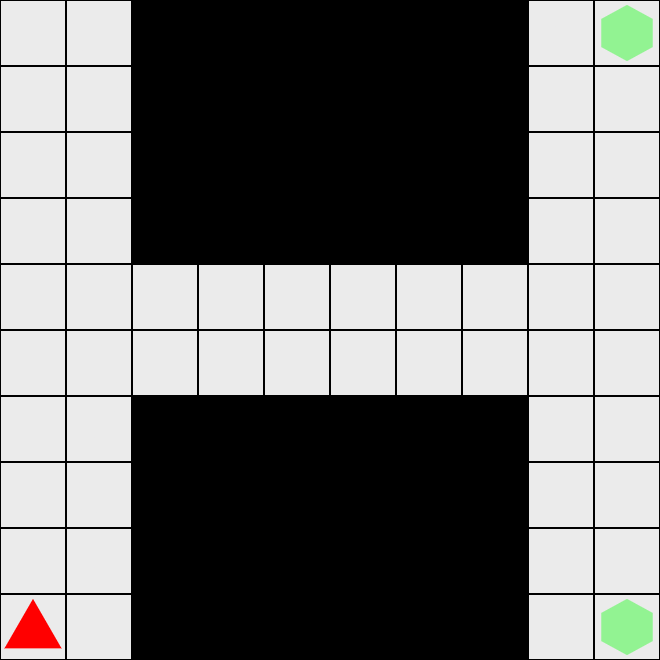}
    \caption{}
    \label{fig:hairpinenv}
    \end{subfigure}
    
    \begin{subfigure}{0.14\textwidth}
    \centering
    \includegraphics[width=0.95\textwidth]{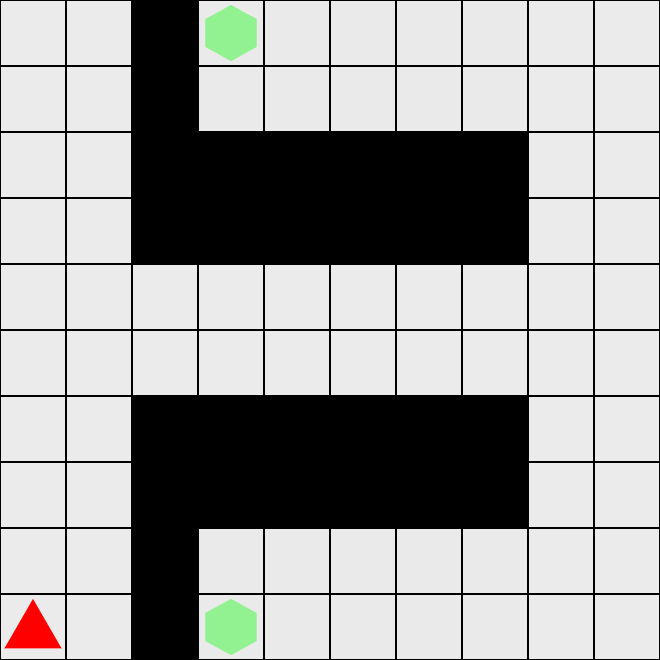}
    \caption{}
    \label{fig:hairpinhardenv}
    \end{subfigure}
    
    \begin{subfigure}{0.14\textwidth}
    \centering
    \includegraphics[width=0.95\textwidth]{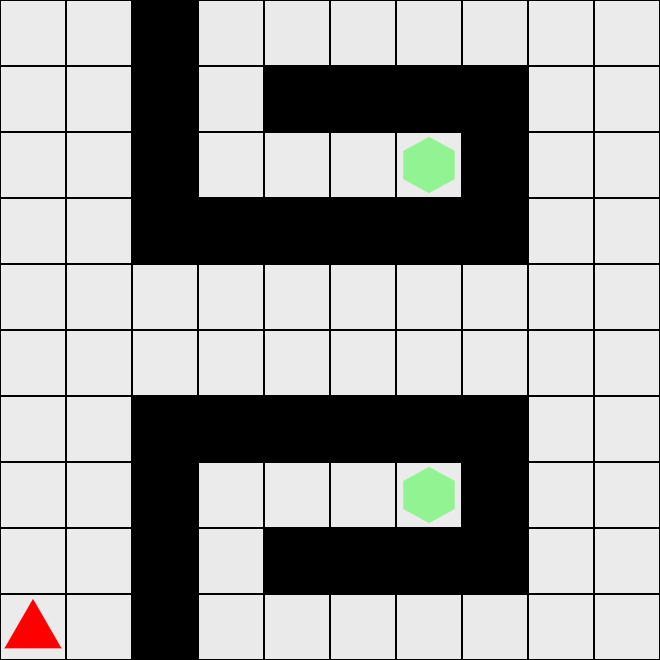}
    \caption{}
    \label{fig:hairpinharderenv}
    \end{subfigure}
    }
    \caption{Three gridworld environments were used in the reward revaluation task \textbf{(a)} Junction \textbf{(b)} Junction Hard \textbf{(c)} Junction Very Hard. Red kites indicate starting locations. White squares are occupiable states and black squared are terminal lava states yielding a $-1$ reward. Green hexagons are potential terminal goal states and yield a $+1$ reward. A single goal is present in each environment at a time. }
    \label{fig:env_diagrams}
\end{figure}

    To assess the reward revaluation capability of the t-SR, we used three tabular gridworld environments which each had two possible goal locations  (Fig \ref{fig:env_diagrams}). Agents learn a policy over 10000 episodes with the reward fixed in one of these locations. The reward is then moved to the other possible location, and the agents must relearn their policy over an additional 10000 episodes. The environments are sparsely rewarded, with $+1$ reward for reaching the terminal goal location, $-1$ for entering one of black \emph{lava} squares (which terminate the episode), and $0$ reward otherwise. 

Five agents were used. A basic Q-learning agent \cite{watkins1992q}, an SR agent\cite{dayan1993improving}, a temporally extended \emph{skip Q-learning} agent \cite{biedenkapp-icml21}, a control SR agent with random temporally extended exploration similar to \cite{dabney2020temporally}, and t-SR. In all cases $\alpha = 1$ was used for value and SR updates, $\gamma = 0.99$ , and $\alpha_r = 1$ was used for updating SR reward vectors. All agents followed an $\epsilon-$greedy exploration policy with $\epsilon = 0.05$. The skip Q and t-SR agents selected both actions and action repetitions in an $\epsilon$-greedy fashion, and used a $j_{max}$ of 7. The SR agent with random temporally extended exploration sampled action repetition randomly with probability $\epsilon$ during training. All results were evaluated under the greedy policy using the same 50 seeds. 

\section{Results}
\begin{figure}[h]
    \centering
    \makebox[\linewidth][c]{
    \begin{subfigure}{0.27\textwidth}
    \centering
    \includegraphics[width=0.95\textwidth]{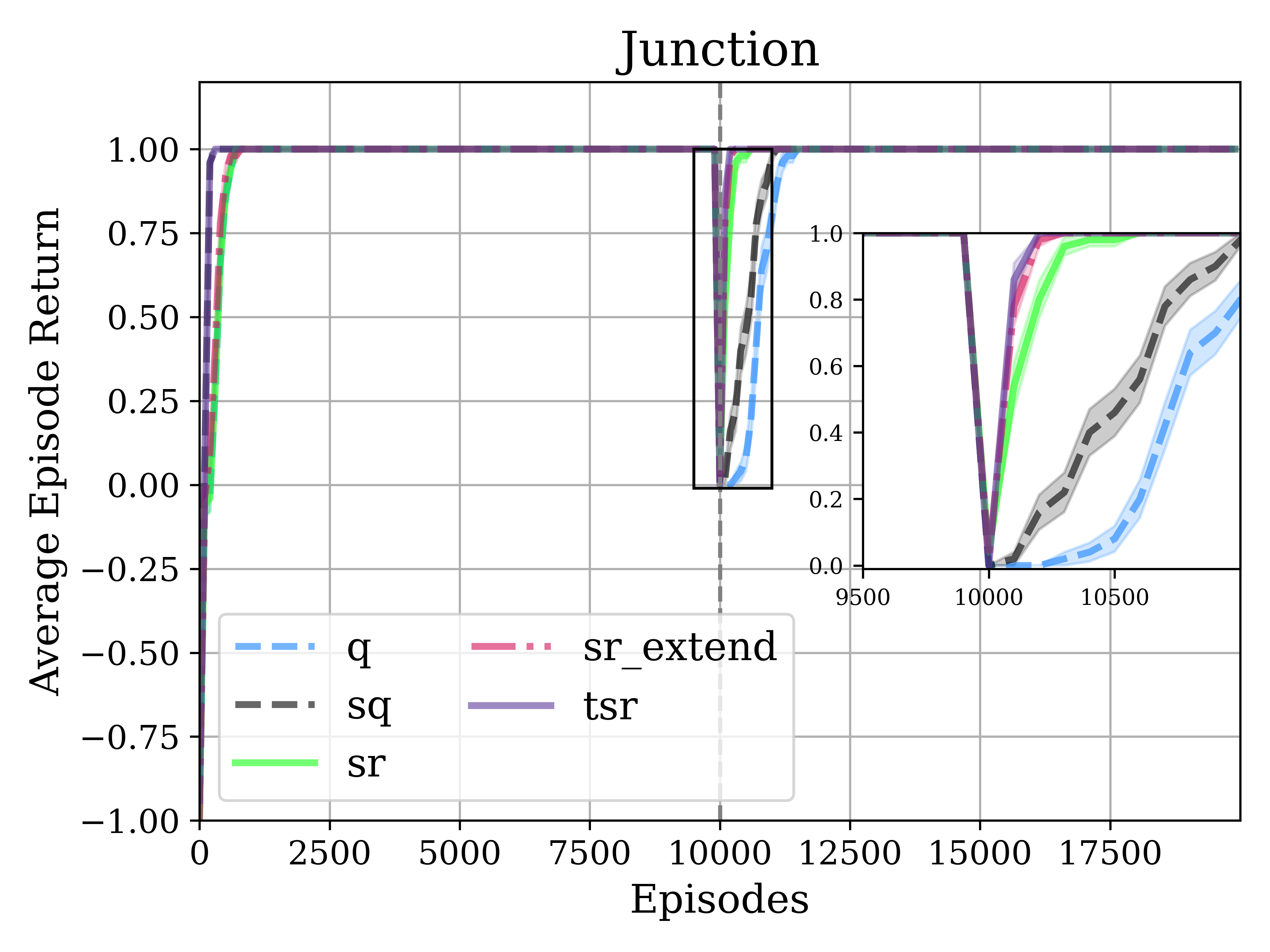}
    \label{fig:reward_hairpin}
    \end{subfigure}
    
    \begin{subfigure}{0.27\textwidth}
    \centering
    \includegraphics[width=0.95\textwidth]{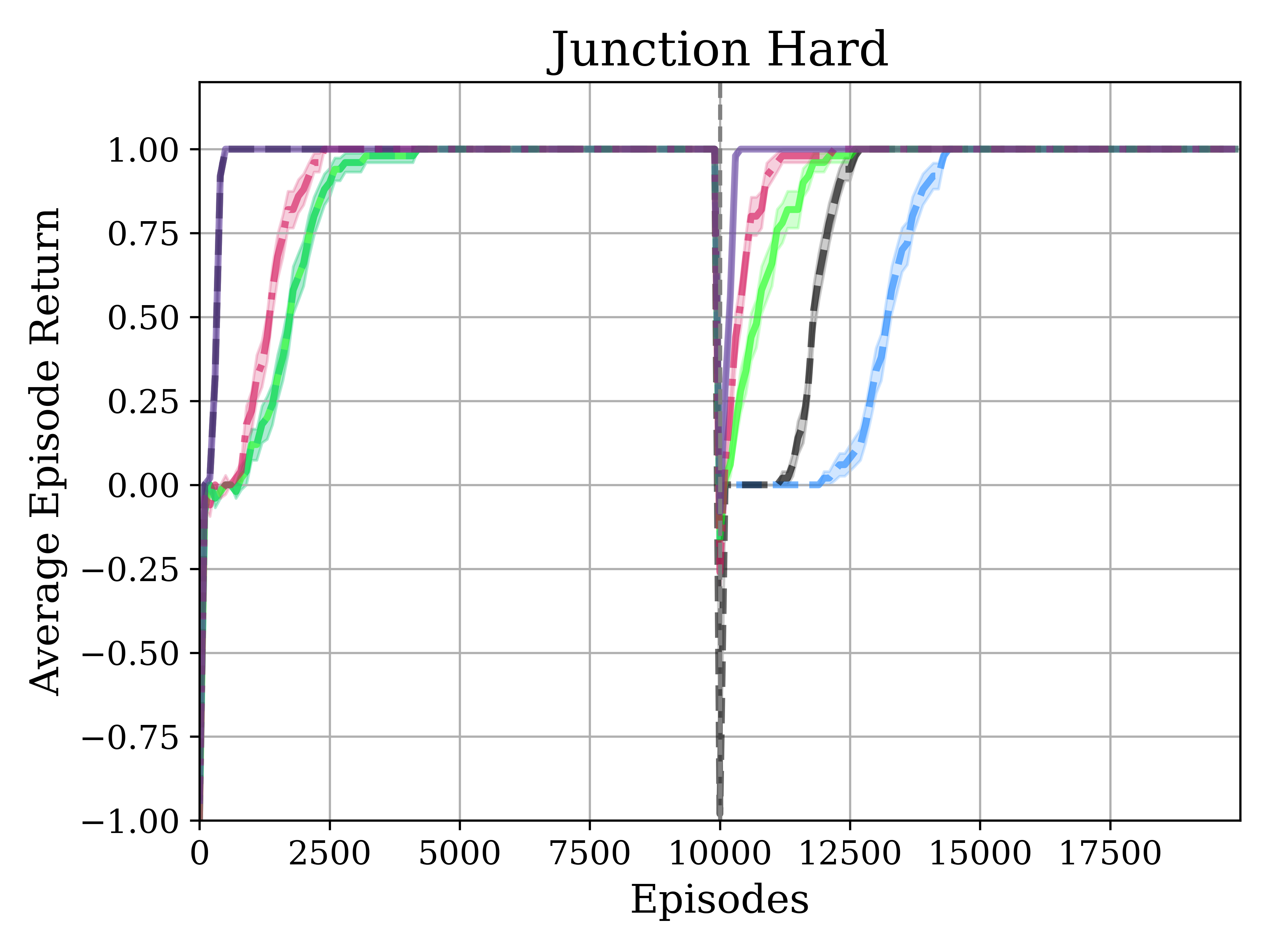}
    \label{fig:reward_hairpinhard}
    \end{subfigure}
    
    \begin{subfigure}{0.27\textwidth}
    \centering
    \includegraphics[width=0.95\textwidth]{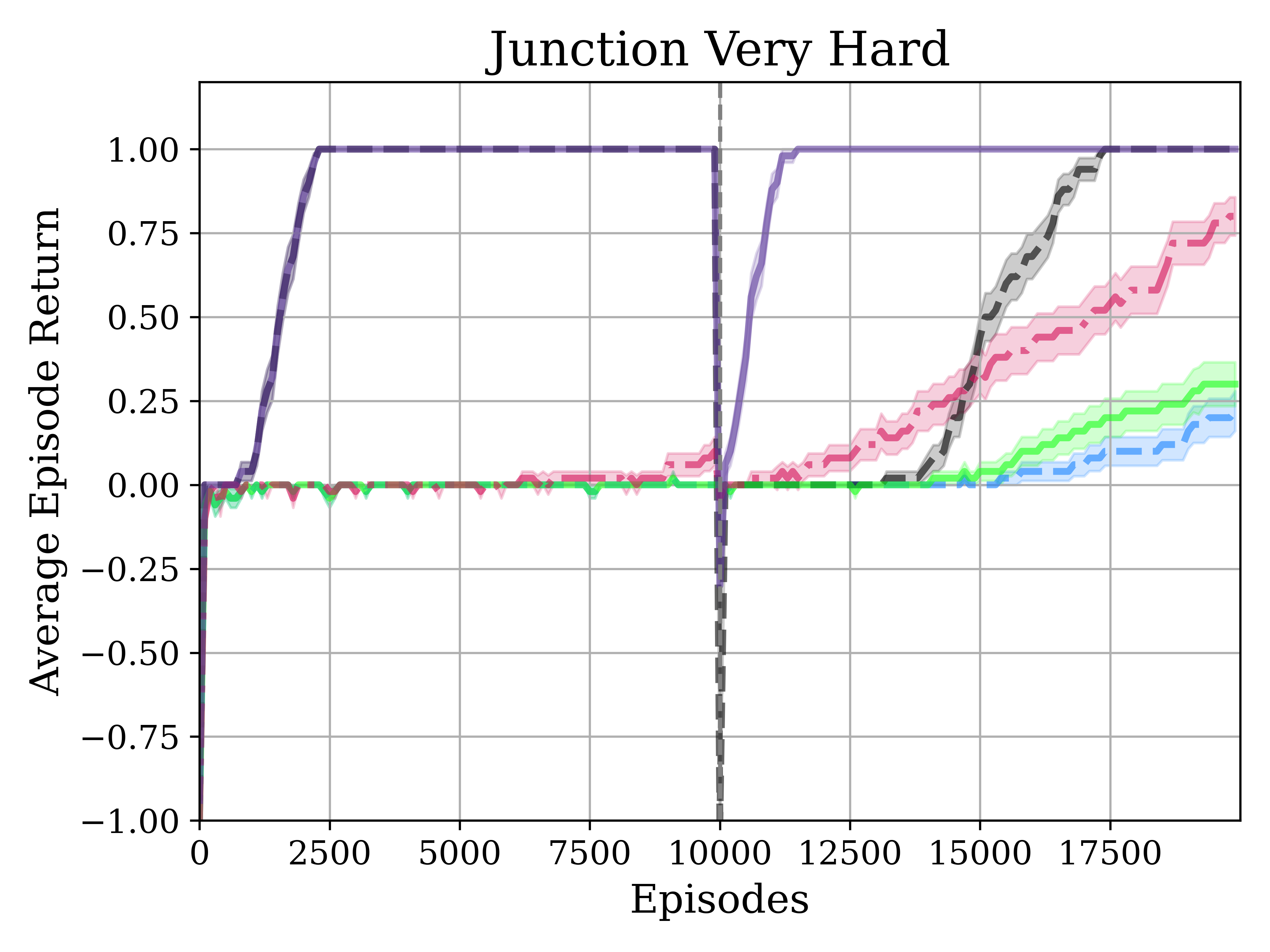}
    \label{fig:reward_hairpinharder}
    \end{subfigure}
    }
   
    \caption{Average episode return under the greedy policy for the three environments measured against the number of episodes trained under the $\epsilon$-greedy policy. Results are plotted as means and standard errors over 50 seeds.}
    \label{fig:rewards}
\end{figure}

Reporting the average episode return for each of the five agents, we see that t-SR relearns an optimal policy to the new goal in fewer episodes than any other agent used (Fig \ref{fig:rewards}). While initially learning at the same rate as skip Q-learning before the reward changes location, t-SR is subsequently able to alter its policy much more rapidly after the change. Note that for the hardest environment, the agents without temporal action repetition policies did not on average find the optimal policy in $10000$ episodes. 

\begin{figure}[h]
    \centering
    \makebox[\linewidth][c]{
    \begin{subfigure}{0.27\textwidth}
    \centering
    \includegraphics[width=0.95\textwidth]{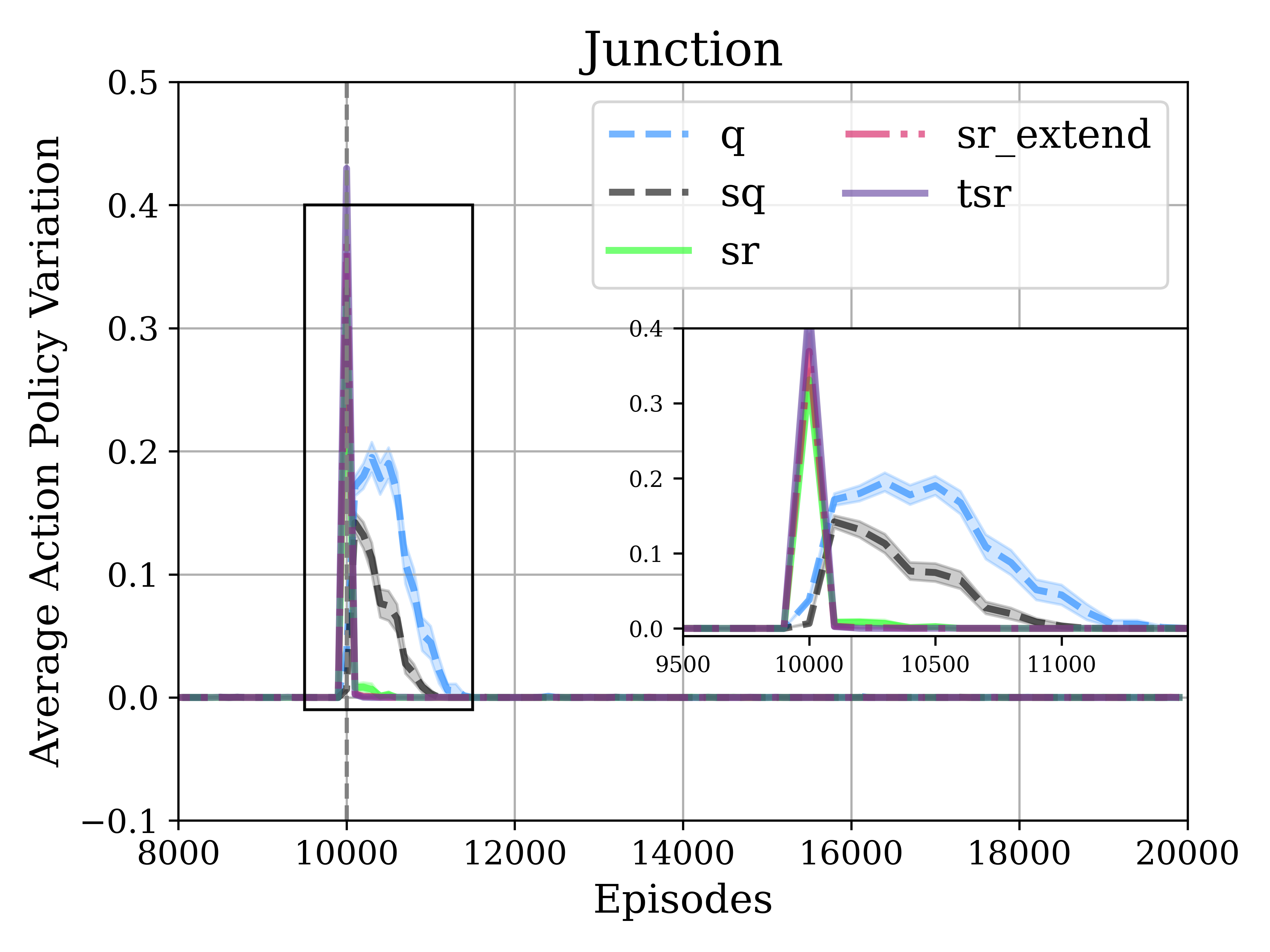}
    \label{fig:tv_hairpin}
    \end{subfigure}
    
    \begin{subfigure}{0.27\textwidth}
    \centering
    \includegraphics[width=0.95\textwidth]{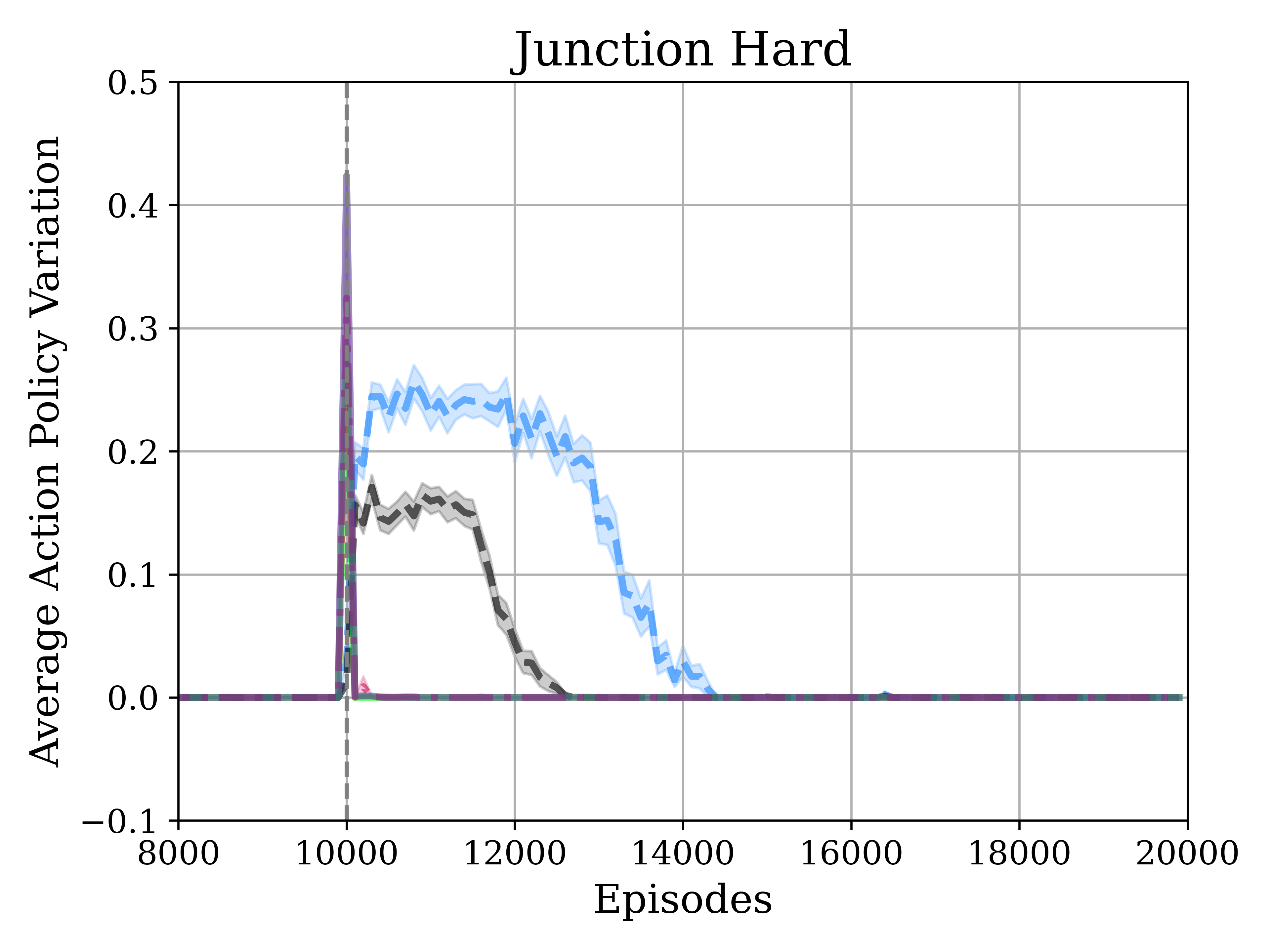}
    \label{fig:tv_hairpinhard}
    \end{subfigure}
    
    \begin{subfigure}{0.27\textwidth}
    \centering
    \includegraphics[width=0.95\textwidth]{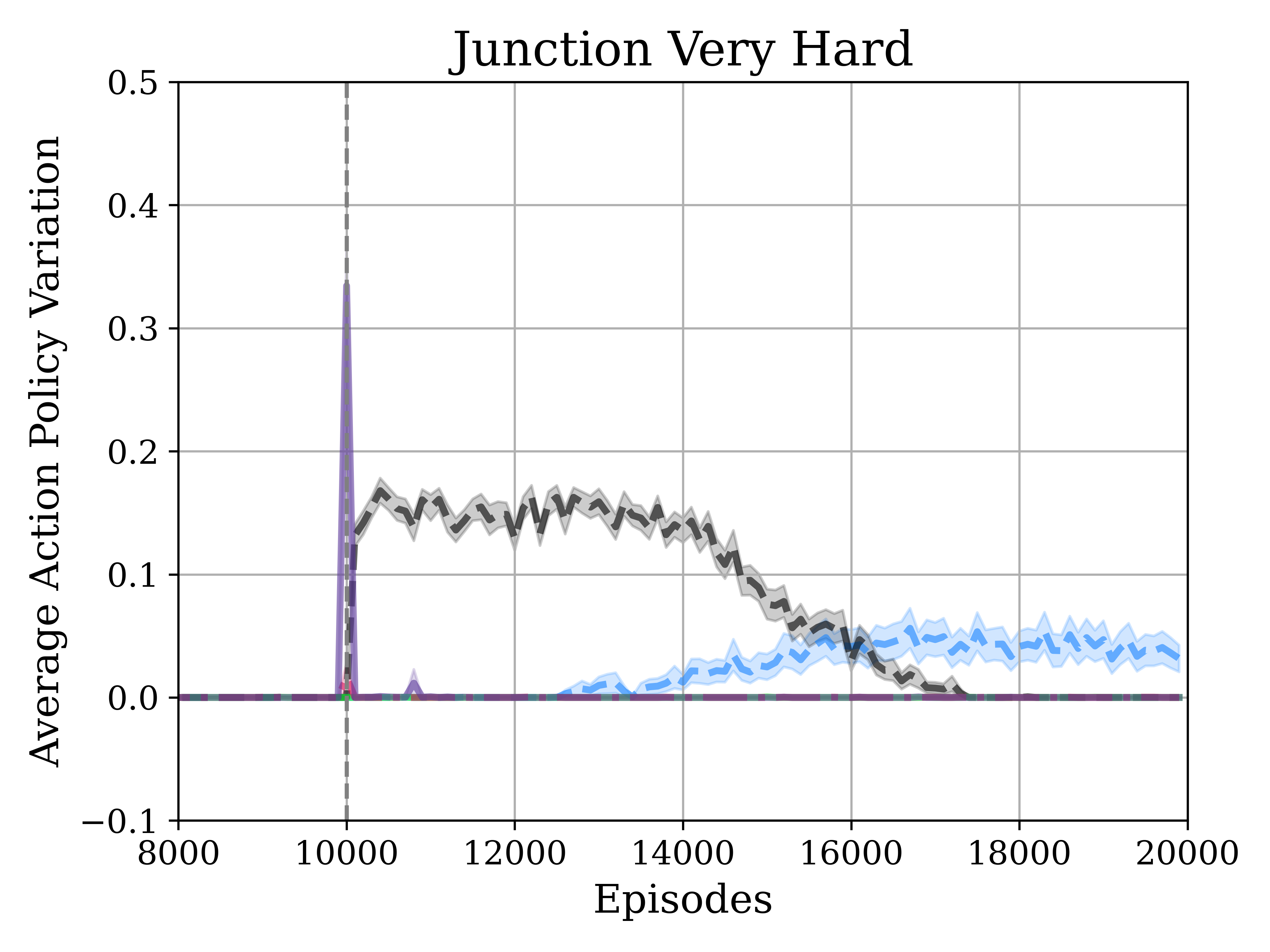}
    \label{fig:tv_hairpinharder}
    \end{subfigure}
    }
    \caption{Average total variation between policies at successive episodes for the three environments measured against the number of episodes trained under the $\epsilon$-greedy policy. Results are plotted as means and standard errors over 50 seeds.}
    \label{fig:TV}
\end{figure}

To assess the rate of policy change after the reward changes, we calculated the average total variation between policies after each episode of learning (Fig \ref{fig:TV}). That is, $TV_\pi(t) =  \frac{1}{|\mathcal{S}|}\sum_{s \in \mathcal{S}}\frac{1}{2}||\pi(s)_{t-1} - \pi(s)_{t} ||_1$. We see that for the SR-based agents, the policy changes sharply and over a much shorter number of episodes compared to the purely model-free agents - which continue to update their policies over a protracted amount of time.

\begin{figure}[h]
    \centering
    \makebox[\linewidth][c]{
    \begin{subfigure}{0.27\textwidth}
    \centering
    \includegraphics[width=0.95\textwidth]{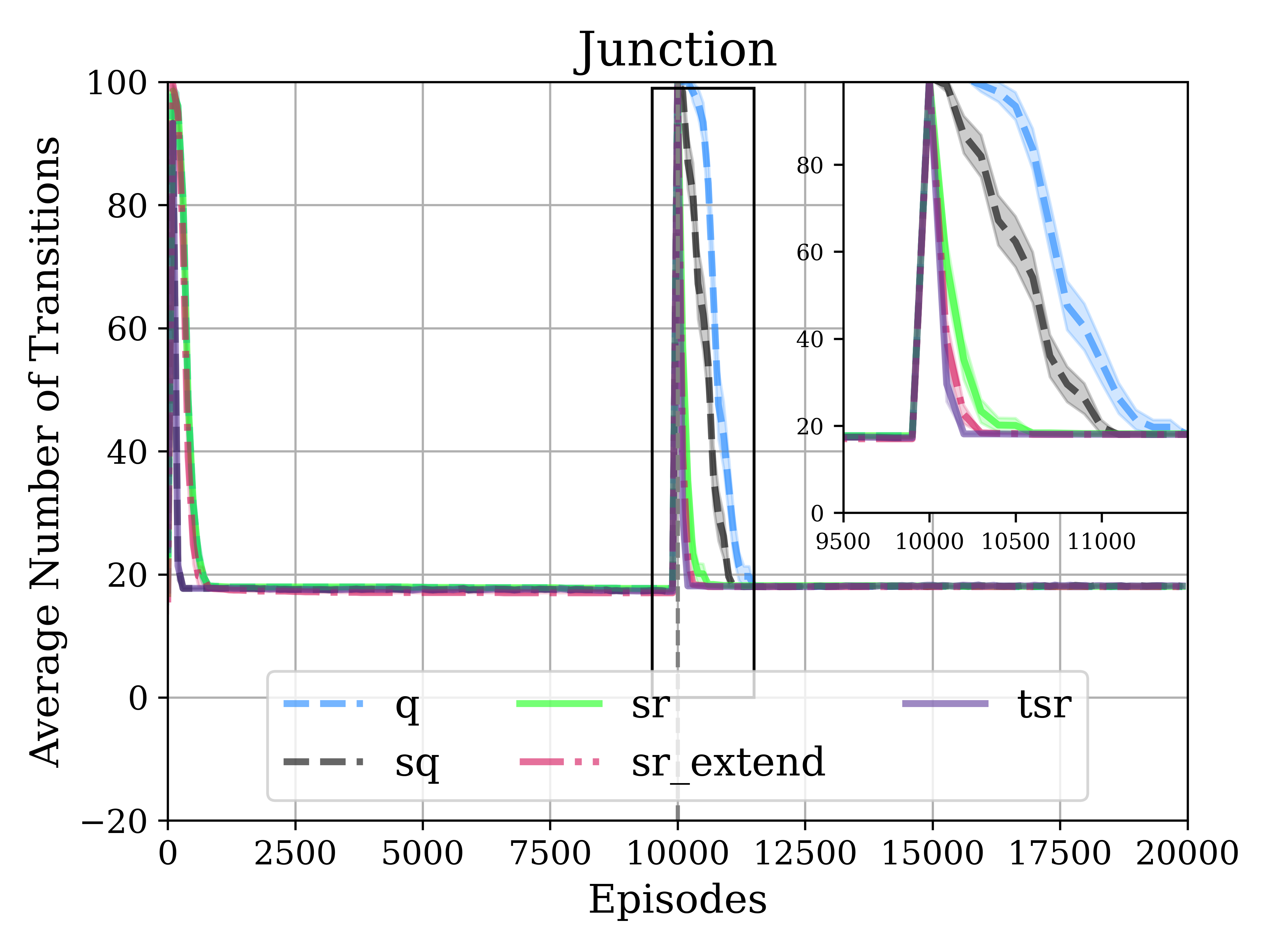}
    \label{fig:steps_hairpin}
    \end{subfigure}
    
    \begin{subfigure}{0.27\textwidth}
    \centering
    \includegraphics[width=0.95\textwidth]{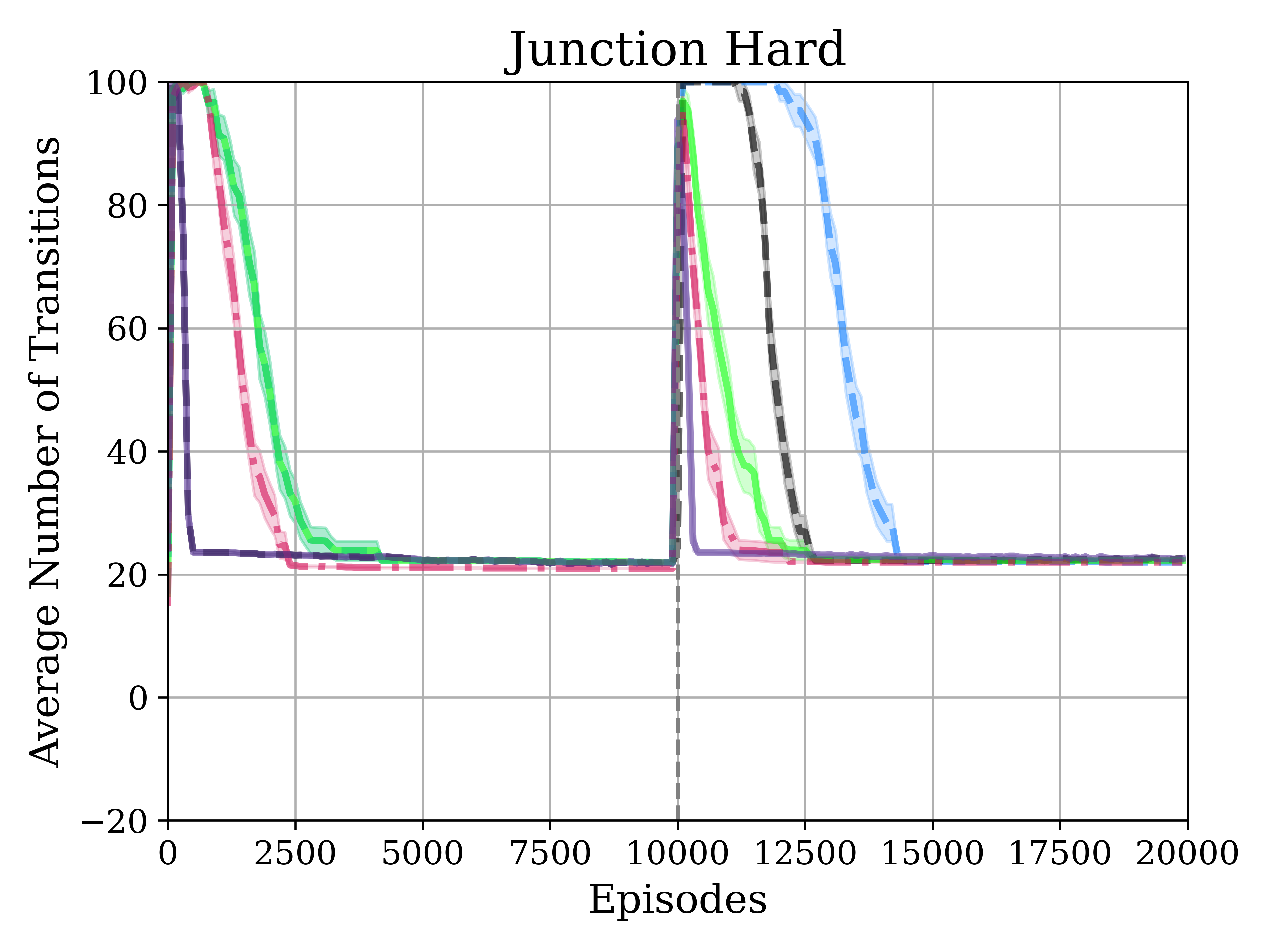}
    \label{fig:steps_hairpinhard}
    \end{subfigure}
    
    \begin{subfigure}{0.27\textwidth}
    \centering
    \includegraphics[width=0.95\textwidth]{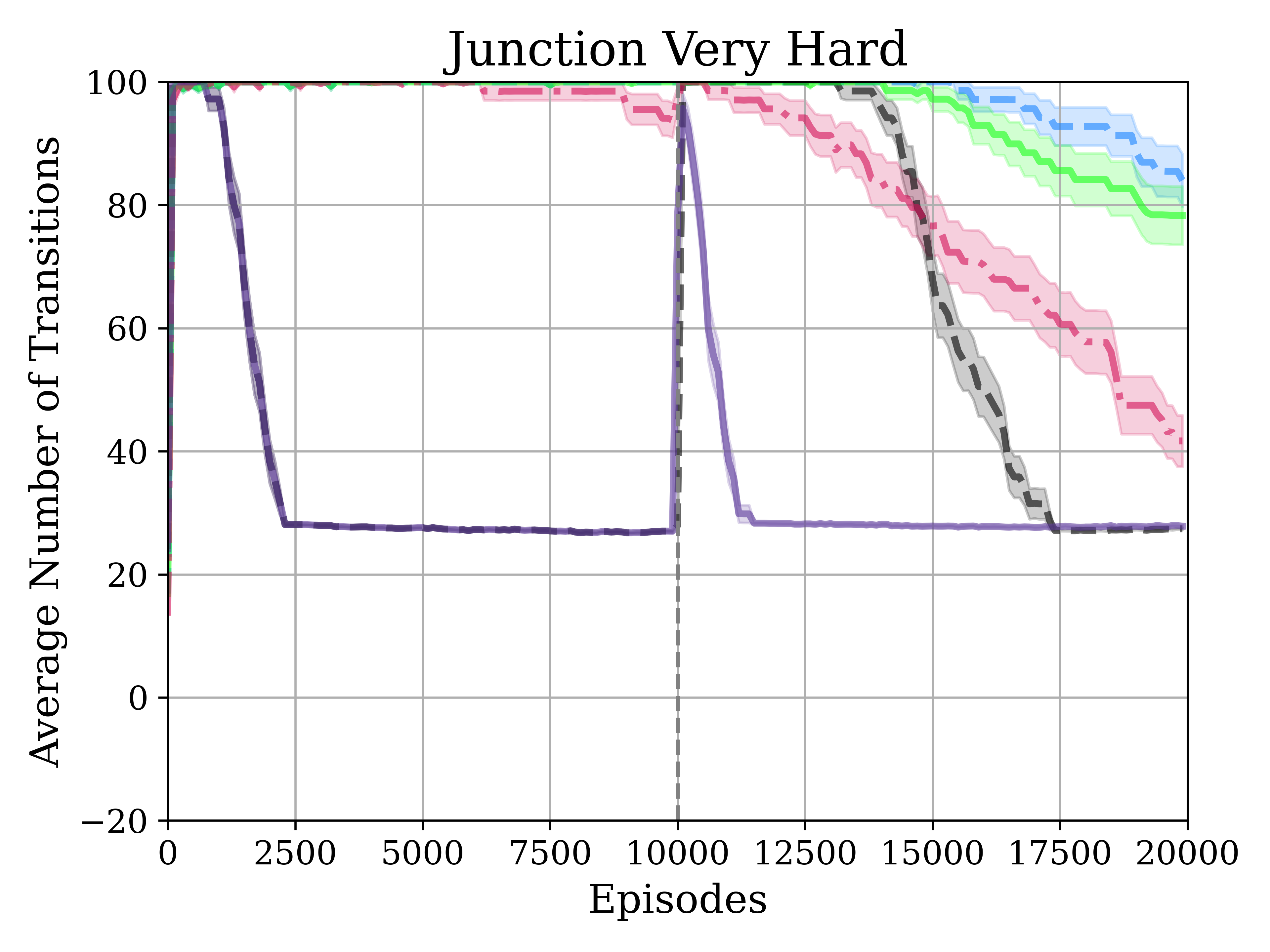}
    \label{fig:stpes_hairpinharder}
    \end{subfigure}
    }

    \centering
    \makebox[\linewidth][c]{
    \begin{subfigure}{0.27\textwidth}
    \centering
    \includegraphics[width=0.95\textwidth]{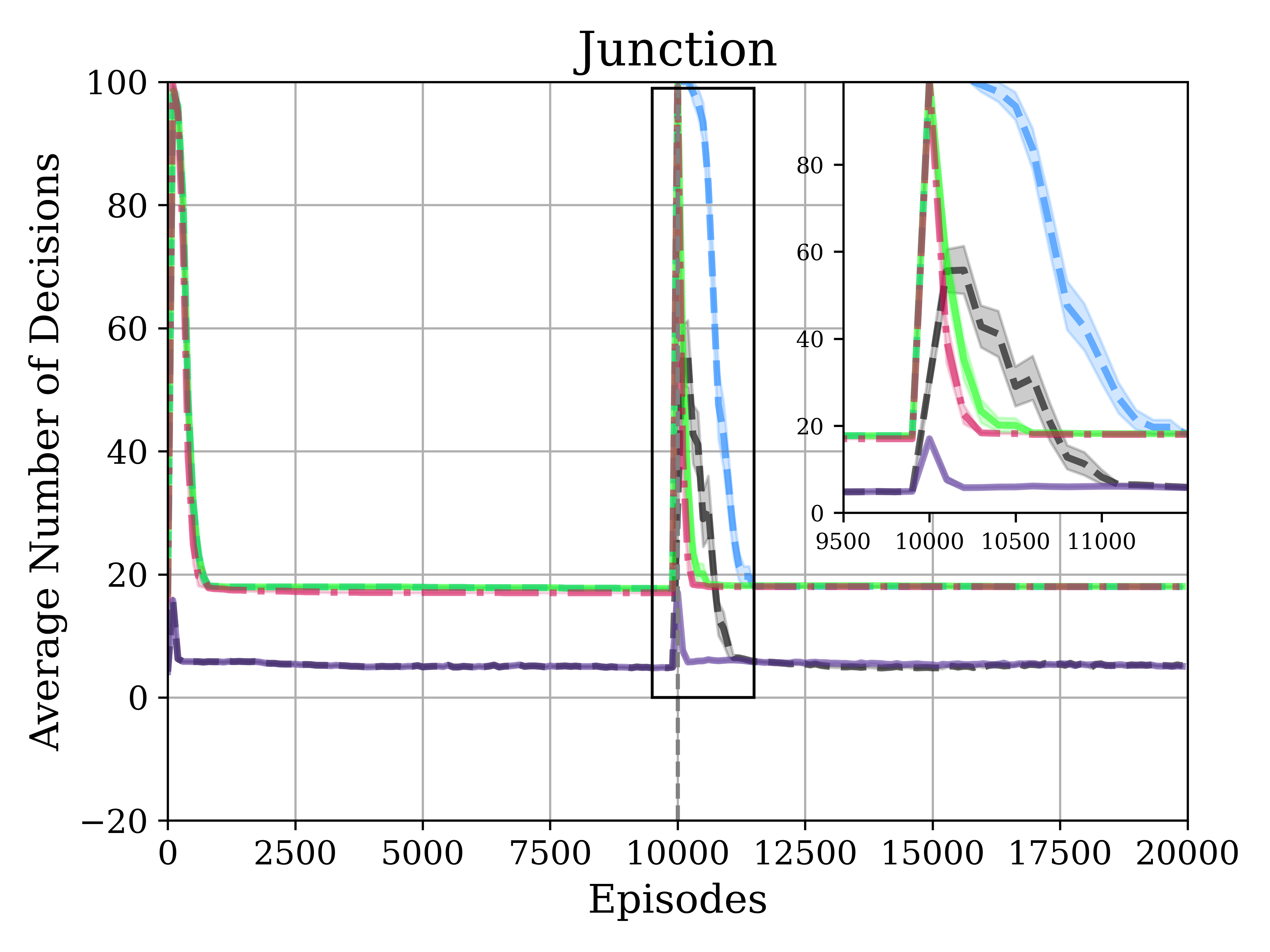}
    \label{fig:steps_hairpin}
    \end{subfigure}
    
    \begin{subfigure}{0.27\textwidth}
    \centering
    \includegraphics[width=0.95\textwidth]{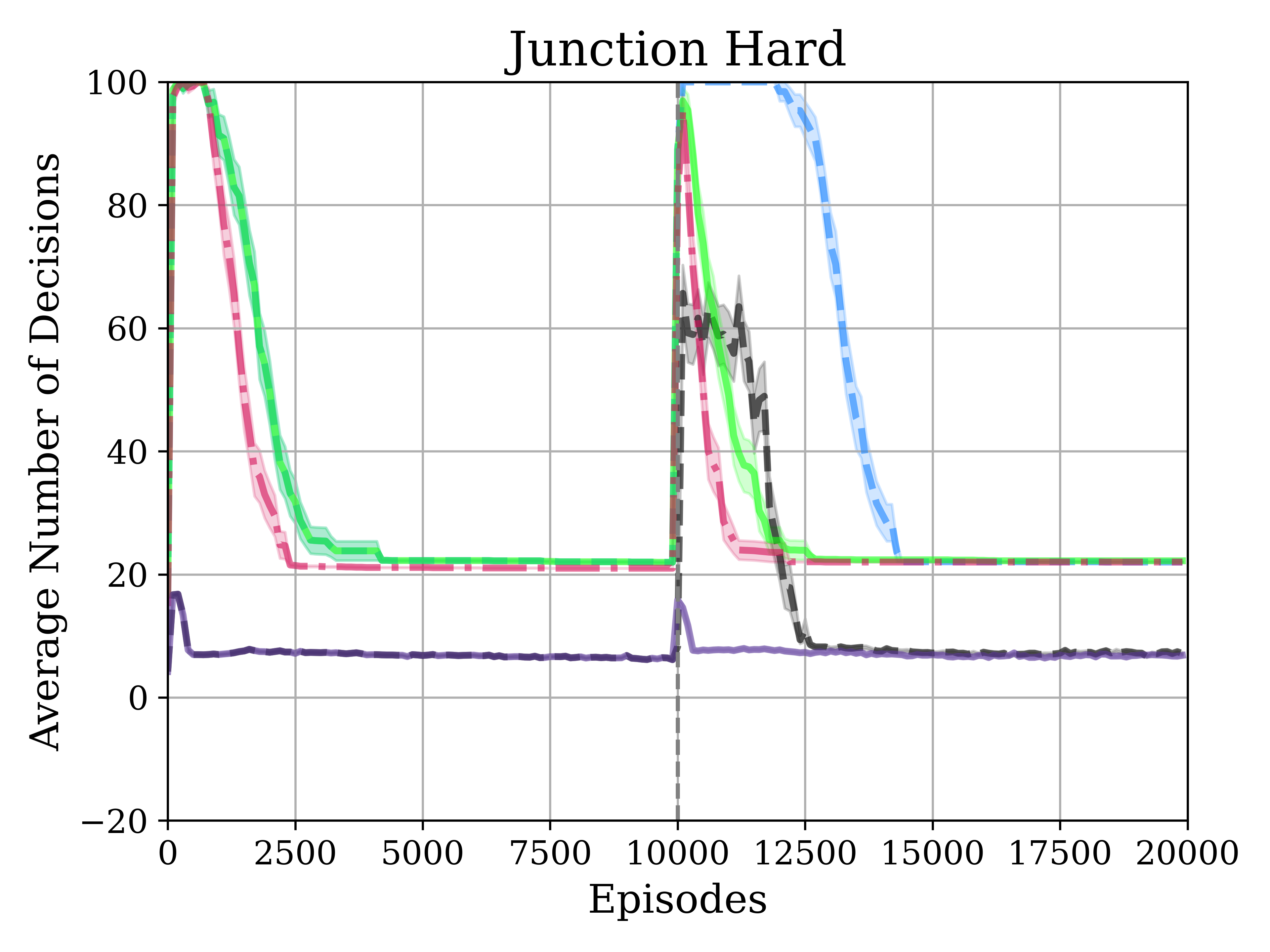}
    \label{fig:steps_hairpinhard}
    \end{subfigure}
    
    \begin{subfigure}{0.27\textwidth}
    \centering
    \includegraphics[width=0.95\textwidth]{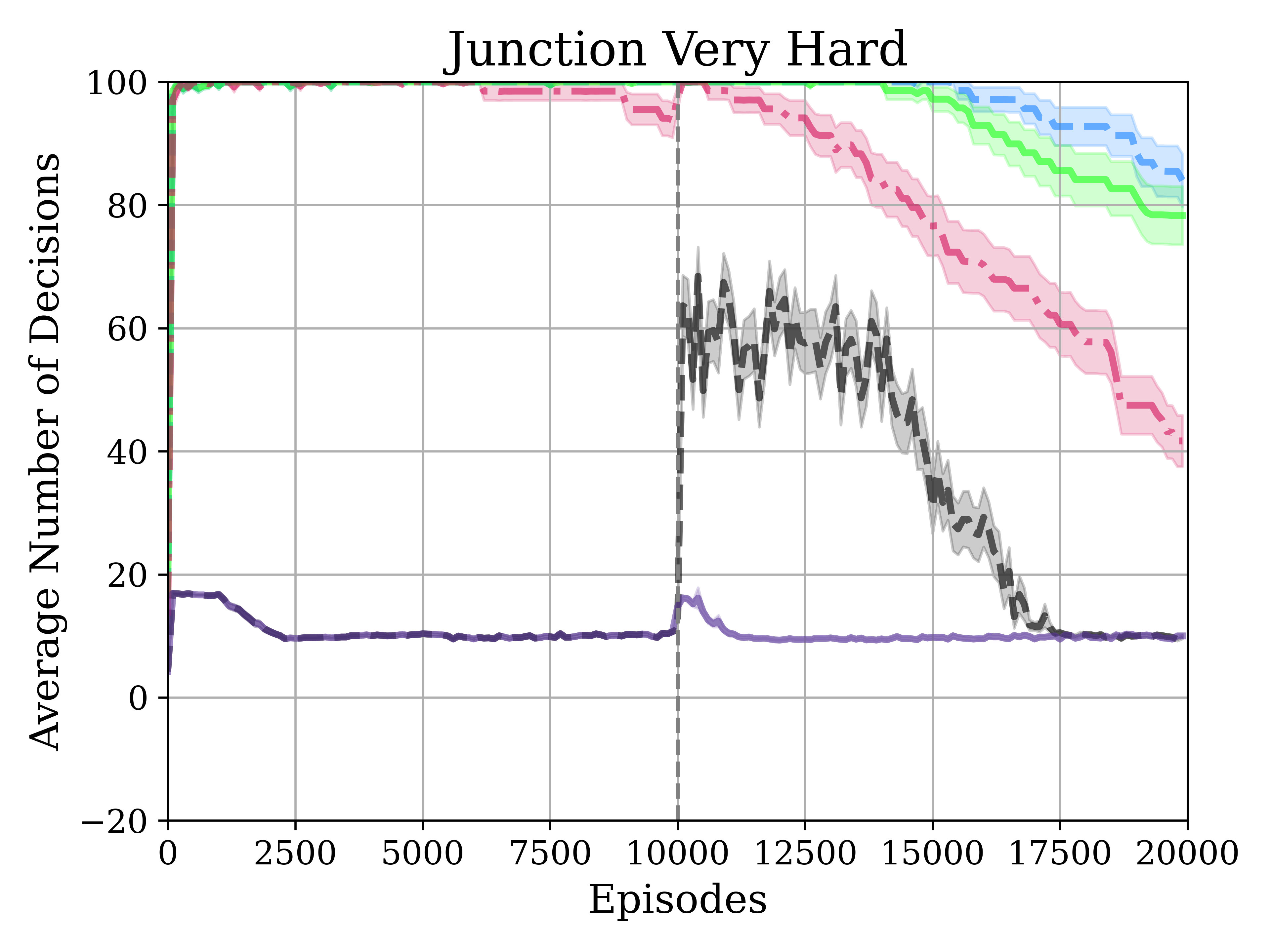}
    \label{fig:steps_hairpinharder}
    \end{subfigure}
    }

    \caption{Average transitions (top row) and decisions (bottom row) performed by agents the three environments measured against the number of episodes trained under the $\epsilon$-greedy policy. \emph{Decisions} refers to how many times the agent queries its action policy during an episode. Results are plotted as means and standard errors over 50 seeds.}
        \label{fig:stepssndlengths}
\end{figure}

Finally, to assess the optimality of the trajectories taken we report statistics on the average path length to reach to goal (Fig \ref{fig:stepssndlengths} top). From the time to convergence to the minimum number of steps needed to solve the environment, we see that t-SR performs reward revaluation approximately 3.5 times faster in the hardest environment than the next fastest agent. Further, t-SR consistently queries its policy a fewer number of times throughout learning (Fig \ref{fig:stepssndlengths} bottom).

\section{Discussion and Future Work}
The success of an agent in performing reward revaluation lies in the ability to discover the new reward. This requires effectively unlearning the current policy being followed. The total variation graphs in Fig \ref{fig:TV} show that, as expected, policies used by SR-based agents change significantly in a short number of episodes compared to purely model-free value based approaches. The improved performance of t-SR over SR or SR with random temporally extended exploration is likely due to the coherent policies used in the episodes immediately after the reward switch. The temporally extended dynamics learnt by t-SR will favour moving significant distances in the state space when compared to SR. This greatly increases the likelihood of following a sequence of actions leading to the new goal. This is also true of skip Q-learning, but these model-free agents are marred by policies that will consistently lead to the old reward location for an extended number of episodes following the reward switch, greatly decreasing the likelihood of finding the new reward. 
We have proposed a new variation of the SR, the t-SR, which shows improved reward revaluation performance compared to SR or other comparable temporally extended methods Our next step is to  extend this framework to the function approximation domain using successor features \cite{barreto2017successor}. We also are currently developing a form of macro action discovery using the t-SR.
\printbibliography[heading=none]

@article{sutton1999between,
  title={Between MDPs and semi-MDPs: A framework for temporal abstraction in reinforcement learning},
  author={Sutton, Richard S and Precup, Doina and Singh, Satinder},
  journal={Artificial intelligence},
  volume={112},
  number={1-2},
  pages={181--211},
  year={1999},
  publisher={Elsevier}
}

@inproceedings{Braylan2015FrameSI,
  title={Frame Skip Is a Powerful Parameter for Learning to Play Atari},
  author={Alexander Braylan and Mark Hollenbeck and Elliot Meyerson and Risto Miikkulainen},
  booktitle={AAAI Workshop},
  year={2015}
}

@misc{hessel2019inductive,
      title={On Inductive Biases in Deep Reinforcement Learning}, 
      author={Matteo Hessel and Hado van Hasselt and Joseph Modayil and David Silver},
      year={2019},
      eprint={1907.02908},
      archivePrefix={arXiv}
}

@inproceedings{lakshminarayanan2017dynamic,
  title={Dynamic action repetition for deep reinforcement learning},
  author={Lakshminarayanan, Aravind and Sharma, Sahil and Ravindran, Balaraman},
  booktitle={AAAI},
  volume={31},
  number={1},
  year={2017}
}

@inproceedings{DBLP:conf/iclr/SharmaLR17,
  author    = {Sahil Sharma and
               Aravind S. Lakshminarayanan and
               Balaraman Ravindran},
  title     = {Learning to Repeat: Fine Grained Action Repetition for Deep Reinforcement
               Learning},
  booktitle = {ICLR},
  year      = {2017},
}

@inproceedings{biedenkapp-icml21,
  author    = {André Biedenkapp and Raghu Rajan and Frank Hutter and Marius Lindauer},
  title     = {{T}empo{RL}: Learning When to Act},
  booktitle = {ICML},
  year = {2021},
  month     = jul,
}

@article{dayan1993improving,
  title={Improving generalization for temporal difference learning: The successor representation},
  author={Dayan, Peter},
  journal={Neural Computation},
  volume={5},
  number={4},
  pages={613--624},
  year={1993},
  publisher={MIT Press}
}

@article{bellemare2013,
   title={The Arcade Learning Environment: An Evaluation Platform for General Agents},
   volume={47},
   journal={Journal of Artificial Intelligence Research},
   publisher={AI Access Foundation},
   author={Bellemare, M. G. and Naddaf, Y. and Veness, J. and Bowling, M.},
   year={2013},
   month={Jun},
   pages={253–279}
}

@article{Momennejad2018PredictingTF,
  title={Predicting the Future with Multi-scale Successor Representations},
  author={Ida Momennejad and Marc W Howard},
  journal={bioRxiv},
  year={2018}
}

@article{watkins1992q,
  title={Q-learning},
  author={Watkins, Christopher JCH and Dayan, Peter},
  journal={Machine Learning},
  volume={8},
  number={3},
  pages={279--292},
  year={1992},
  publisher={Springer}
}

@inproceedings{dabney2020temporally,
  title={Temporally-Extended $\varepsilon$-Greedy Exploration},
  author={Dabney, Will and Ostrovski, Georg and Barreto, Andre},
  booktitle={ICLR},
  year={2020}
}

@article{barreto2017successor,
  title={Successor features for transfer in reinforcement learning},
  author={Barreto, Andr{\'e} and Dabney, Will and Munos, R{\'e}mi and Hunt, Jonathan J and Schaul, Tom and van Hasselt, Hado P and Silver, David},
  journal={NeurIPS},
  volume={30},
  year={2017}
}

@article{pateria2021hierarchical,
  title={Hierarchical reinforcement learning: A comprehensive survey},
  author={Pateria, Shubham and Subagdja, Budhitama and Tan, Ah-hwee and Quek, Chai},
  journal={CSUR},
  volume={54},
  number={5},
  pages={1--35},
  year={2021},
  publisher={ACM New York, NY, USA}
}

@article{russek2017predictive,
  title={Predictive representations can link model-based reinforcement learning to model-free mechanisms},
  author={Russek, Evan M and Momennejad, Ida and Botvinick, Matthew M and Gershman, Samuel J and Daw, Nathaniel D},
  journal={PLoS computational biology},
  volume={13},
  number={9},
  pages={e1005768},
  year={2017},
  publisher={Public Library of Science San Francisco, CA USA}
}

\end{document}